\documentclass{MITstyle}

\usepackage{chngcntr}
\usepackage{hyperref}
\usepackage{microtype}

\title{A Lightweight and Extensible Cell Segmentation and Classification Model for Whole Slide Images}

\author{Nikita Shvetsov~$^{1, }$\footnote{Correspondence e-mail: nikita.shvetsov@uit.no}, Thomas K. Kilvaer~$^{2, 3}$, Masoud Tafavvoghi~$^{4}$, Anders Sildnes~$^{1}$, \\ Kajsa Møllersen~$^{4}$, Lill-Tove Rasmussen Busund~$^{5, 6}$, Lars Ailo Bongo~$^{1}$ \\
\vspace{1em} 
\normalfont{\small $^{1}$Department of Computer Science, UiT The Arctic University of Norway}\\
\normalfont{\small $^{2}$Department of Oncology, University Hospital of North Norway}\\
\normalfont{\small $^{3}$Department of Clinical Medicine, UiT The Arctic University of Norway}\\
\normalfont{\small $^{4}$Department of Community Medicine, UiT The Arctic University of Norway}\\
\normalfont{\small $^{5}$Department of Medical Biology, UiT The Arctic University of Norway} \\
\normalfont{\small $^{6}$Department of Clinical Pathology, University Hospital of North Norway} 
}

\begin{document}
\maketitle

\section*{Abstract}

\begin{abstract}
Developing clinically useful cell-level analysis tools in digital pathology remains challenging due to limitations in dataset granularity, inconsistent annotations, high computational demands, and difficulties integrating new technologies into workflows. To address these issues, we propose a solution that enhances data quality, model performance, and usability by creating a lightweight, extensible cell segmentation and classification model. 

First, we update data labels through cross-relabeling to refine annotations of PanNuke and MoNuSAC, producing a unified dataset with seven distinct cell types. Second, we leverage the H-Optimus foundation model as a fixed encoder to improve feature representation for simultaneous segmentation and classification tasks. Third, to address foundation models' computational demands, we distill knowledge to reduce model size and complexity while maintaining comparable performance. Finally, we integrate the distilled model into QuPath, a widely used open-source digital pathology platform. 

Results demonstrate improved segmentation and classification performance using the H-Optimus-based model compared to a CNN-based model. Specifically, average $R^2$ improved from 0.575 to 0.871, and average $PQ$ score improved from 0.450 to 0.492, indicating better alignment with actual cell counts and enhanced segmentation quality. The distilled model maintains comparable performance while reducing parameter count by a factor of 48. By reducing computational complexity and integrating into workflows, this approach may significantly impact diagnostics, reduce pathologist workload, and improve outcomes. Although the method shows promise, extensive validation is necessary prior to clinical deployment.
\end{abstract}
\clearpage

\section{Introduction}
In digital pathology, accurate segmentation and classification of cells are crucial for many diagnostic, prognostic, and predictive analyses \cite{Jaber_Beziaeva_etal._2019,Lin_Pan_etal._2022,Park_Ock_etal._2022,Shen_Choi_etal._2024}. Nowadays, developments in computational pathology offer multiple solutions \cite{H._Qu_P._Wu_etal._2020,Javed_Mahmood_etal._2020} to utilize cell-level datasets to train machine learning models that solve these problems. The quality and specificity of training datasets are critical for robust and accurate models. Adhering to the principle of "garbage in, garbage out", it is essential to ensure that these datasets are extensively and accurately labeled with distinct classes that reflect the diverse biological characteristics of different cell types. Unfortunately, the number of open-source datasets comprising such high-quality annotations is limited. Existing cell segmentation datasets \cite{Gamper_Koohbanani_etal._2019,Graham_Vu_etal._2019,Verma_Kumar_etal._2021} may offer extensive annotations for certain cell types while providing more general labels for others. For example, in PanNuke, which is one of the largest open-source datasets comprising labeled cells, various types of morphologically and functionally different inflammatory cells like macrophages and lymphocytes are clustered in a broad "inflammatory" class. Consequently, these classes are frequently omitted from analyses or aggregated into broader meta-classes \cite{Gamper_Koohbanani_etal._2020} and likely interfere with other cell classes included in the dataset. This and similar inconsistencies in annotation granularity limit the ability of machine learning models to learn the comprehensive and nuanced features necessary for accurate cell segmentation and classification. To address these challenges, methods for refining and standardizing dataset annotations are essential to enhance the quality of training data.

A complementary approach to mitigate the absence of high-quality training data is the use of foundation models. Foundation models as encoders are defined as large-scale, versatile networks pre-trained on vast, diverse datasets using self-supervised learning, contrasting with convolutional neural network (CNN) pre-trained encoders that rely on supervised learning with labeled data. In practice, foundation models leverage enormous amounts of weakly or unlabeled data from millions of whole slide images (WSIs) and employ self-attention mechanisms to capture long-range dependencies and global context \cite{Chen_Ding_etal._2024,Saillard_Jenatton_etal._2024,Vorontsov_Bozkurt_etal._2024,Xu_Usuyama_etal._2024}. As a consequence, foundation models are able to produce transferable feature representations across different cell types and tissue environments. The feature representations can be leveraged by decoder networks to produce segmentation masks and pixel-level classifications. Because foundation models have comprehensive feature representations, they can be effectively fine-tuned using much smaller amounts of cell-level data compared to the large datasets needed to train models from scratch. Furthermore, foundation models incorporate adversarial training elements or contrastive learning \cite{Chen_Ding_etal._2024,Xu_Usuyama_etal._2024}, enhancing their resilience and adaptability by exposing them to challenging and varied scenarios during training. This may result in more generalizable models, often making them well-suited for diverse and complex tasks in digital pathology.

Despite the inherent advantages of foundation models, their deployment for practical use faces its own obstacles. In particular, they require substantial computational power, financial investments and rigorous testing to ensure reliability and efficacy for a given task \cite{Akkus_Dangott_etal._2022,Dragomir_Cocuz_etal._2022,Go_2022,Jafri_Farooqui_etal._2024}. Moreover, while foundation models enhance feature representation and performance, they depend on the quality of available annotations for decoder fine-tuning and, like any other model, cannot resolve existing inconsistencies or ambiguities in data labels. Therefore, there remains a critical need for solutions that address both data quality and practical deployment considerations.
Further, integrating new technologies into existing clinical workflows often encounters resistance, as it necessitates adjustments to established diagnostic processes. So, there is a need to develop solutions that could be integrated into current practices, minimizing the burden on medical professionals to adopt new tools \cite{King_Williams_etal._2023}.

\begin{figure}[h!]
    \centering
    \includegraphics[width=\textwidth, height=0.82\textheight, keepaspectratio]{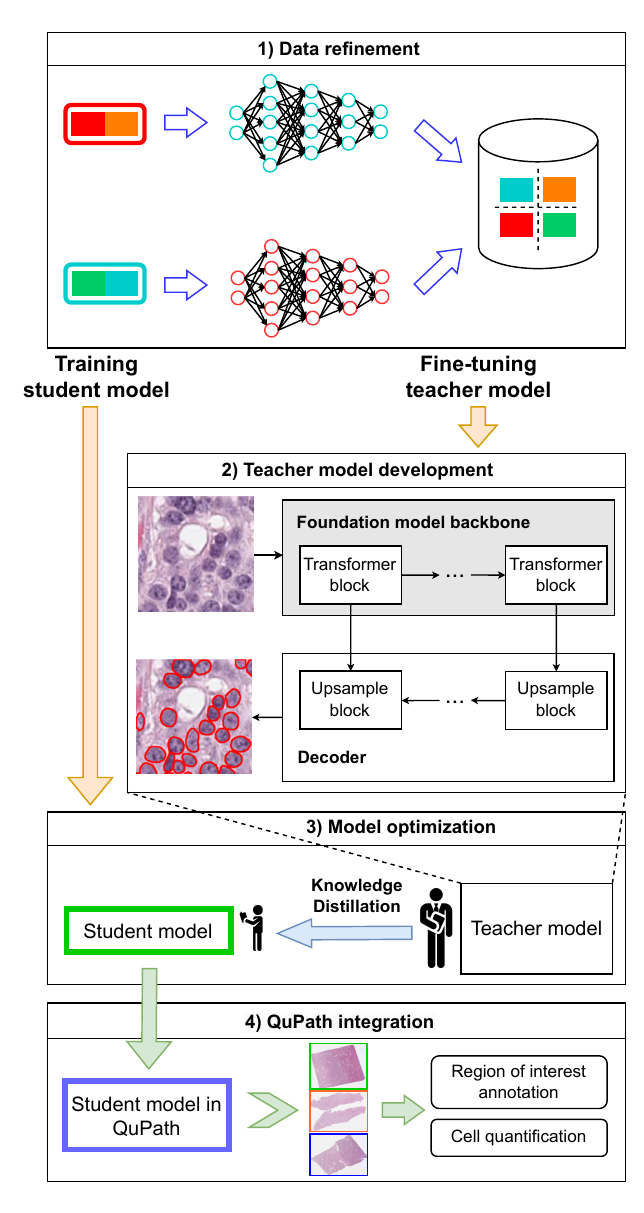}
    \caption{Overview of the proposed solution, including 1) Data refinement using cross-relabeling, 2) Teacher model development and fine tuning, 3) Student model optimization with knowledge distillation and 4) Student model and QuPath integration}
    \label{fig:fig1}
\end{figure}

Existing solutions \cite{Goldsborough_Philps_etal._2024,Hörst_Rempe_etal._2024}, while addressing some aspects of these challenges, fall short in providing a comprehensive approach. To address the data quality and clinical deployment issues, we propose a multi-faceted solution that encompasses data refinement, model optimization, and integration with existing pathology tools (\hyperref[fig:fig1]{Figure 1}). The outcome is a lightweight cell segmentation and classification model that can be integrated into digital pathology workflows for practical clinical use.

\clearpage

Our approach begins with preparing the data for the fine-tuning and training of the machine learning models. We create a refined dataset, acquired via cross-relabeling two cell-level datasets, enhancing annotation specificity and consistency of the labeled data. Subsequently, we create a cell segmentation and classification model based on the foundation model. We leverage the foundation model as a fixed encoder and fine-tune a decoder using the refined dataset to improve generalization across diverse tissue- and cell types.
To ensure that the model remains lightweight and deployable in a possibly resource-constrained environment, we employ knowledge distillation to approximate the functionality of the foundation model. Finally, to facilitate the practical application of our model in digital pathology workflows, we integrate it with the QuPath \cite{Bankhead_Loughrey_etal._2017} application. Each methodological component contributes to the overarching goal of enhancing model performance, generalizability, and usability in clinical settings.

The primary contributions of this paper are:
\begin{enumerate}
    \item \textit{Data labels refinement through cross-relabeling:}
    
    We propose a new method for refining labels of cell-level datasets through cross-relabeling. This method employs classification models to re-label broad and ambiguous instances, resulting in a more diverse dataset. Our evaluation demonstrates that these classification models achieve high accuracy on test subsets, indicating the reliability of the method for label refinement.

    \item \textit{Enhanced model performance via foundation models:}
    
    We employ a foundation model as a feature extractor for the cell segmentation and classification task. In comparison with training a CNN model from scratch, the foundation model backbone only needs fine-tuning, which significantly reduces training time, computational resources and data requirements. We show that using a foundation model encoder leads to better performance in cell segmentation and classification networks than using a CNN-based encoder. This improvement may enable the model to generalize more effectively across various tissue types and imaging methods.
    
    \item \textit{Model optimization through knowledge distillation:}
    
    We show that a smaller student model trained using knowledge distillation on the refined dataset obtained via our cross-relabeling approach from a foundation model achieves comparable performance in cell segmentation and quantification tasks. As a result, this model is more suitable for deployment in environments without high-performance computing resources.
    
    \item \textit{Integration with QuPath:}
    
    We integrate the distilled cell segmentation and classification model into QuPath, a widely used open-source digital pathology platform, to accelerate clinical adaptation by enabling pathologists to more easily incorporate advanced computational tools into their existing workflows.
\end{enumerate}

Through these methodological steps, we aim to bridge the gap between advanced machine learning techniques and practical clinical applications, making accurate and efficient digital pathology accessible in a broader range of healthcare settings.

\section{Refining Existing Datasets Using Cross-Relabeling}
\begin{figure}[h!]
    \centering
    \includegraphics[width=\textwidth]{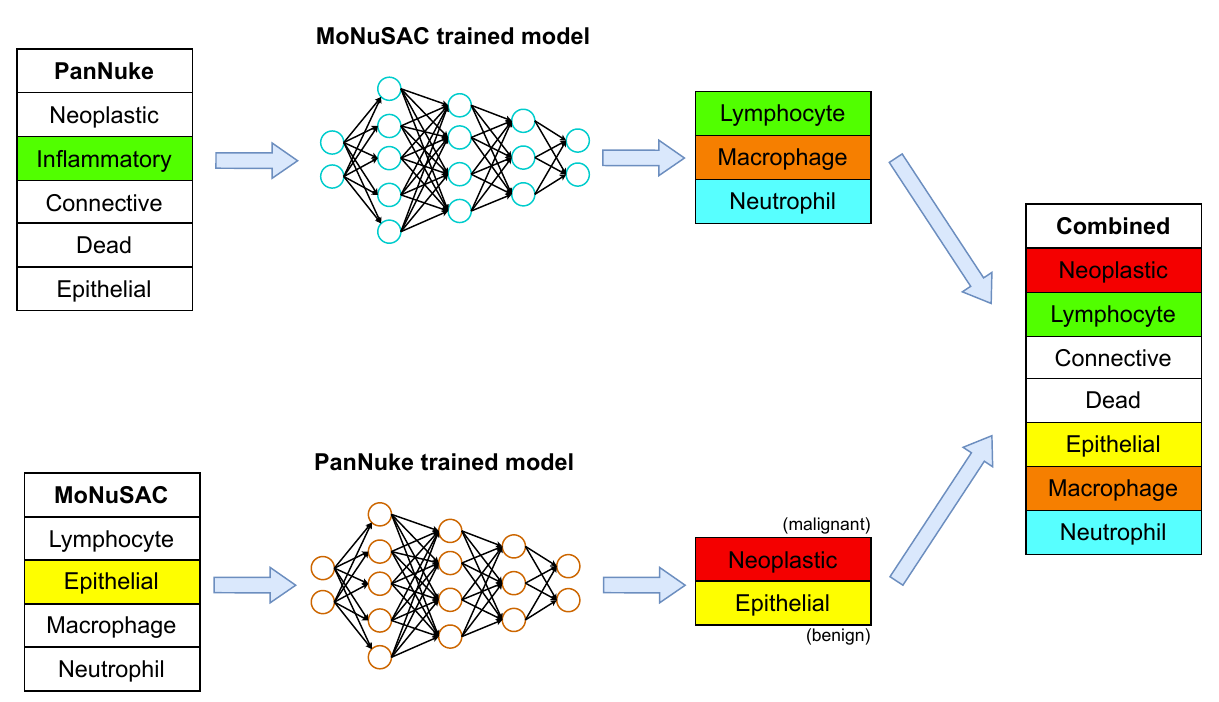}
    \caption{Refined dataset generation via cross relabeling}
    \label{fig:fig2}
\end{figure}

To address the limitations of sparse and ambiguous labeling of cell-level datasets, we propose a generalizable cross-relabeling strategy that can be applied to any dataset containing broadly categorized or imprecisely labeled cell types. This approach involves training and subsequently leveraging classification models to refine broad categories into more specific or biologically relevant classes.
When applied to cell-level data, the methodology includes extracting individual cell images from the dataset patches, preprocessing these images to standardize the size and accommodate partial cells, and then training deep learning classifiers capable of distinguishing between the finer cell subtypes within the coarser categories. 
To illustrate our approach, we focus on the PanNuke \cite{Gamper_Koohbanani_etal._2020, Gamper_Koohbanani_etal._2019} and MoNuSAC \cite{Verma_Kumar_etal._2021} datasets that we have used to train models for cell quantification in our previous works \cite{Shvetsov_Grønnesby_etal._2022,Shvetsov_Sildnes_etal._2024}. We find that for better cell differentiation we have to introduce more granular labels. PanNuke includes a broad classification of "inflammatory" cells, encompassing lymphocytes, macrophages, and neutrophils. Each cell type differs significantly in structure, function, and clinical relevance. Conversely, MoNuSAC uses the label "epithelial" for a class that comprises both benign epithelial cells and malignant neoplastic cells. This practice makes it challenging to differentiate between benign and malignant epithelial cells in the dataset, which is a critical distinction when identifying tumor areas within tissue samples. To address these issues, we implement a cross-relabeling strategy as shown in \hyperref[fig:fig2]{Figure 2}. The key components are two classification models: one is trained on singular cell images from PanNuke data to classify the epithelial meta-class into epithelial and neoplastic classes. The other is trained on MoNuSAC to refine the inflammatory class into lymphocytes, neutrophils, and macrophages.

\begin{figure}[h!]
    \centering
    \includegraphics[width=\textwidth]{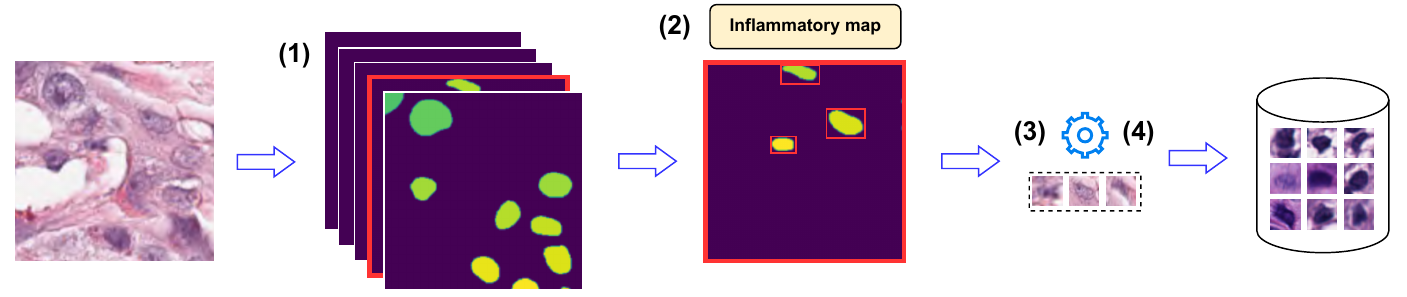}
    \caption{Cell instances preprocessing including (1) cell map extraction, (2) bounding box delineation, (3) adjusting cell boxes and (4) cropping and resizing of cell images}
    \label{fig:fig3}
\end{figure}

The refining approach consists of three consecutive steps. The first is the preprocessing step, in which we extract individual cells from both datasets (\hyperref[fig:fig3]{Figure 3}). The specifics of PanNuke and MoNuSAC patch preparation before cell preprocessing are provided in \hyperref[chap:S1]{Appendix S1}.

During preprocessing, we extract cell type maps from the ground truth label mask and calculate bounding boxes around each cell instance. To accommodate partial cells at patch borders, a common issue in cropped patch images, we employ mirror padding and extend the field of view of the cell label by 15 pixels to capture adjacent cells. We then crop and resize the identified regions to $64 \times 64$ pixels using bicubic interpolation.

The preprocessed PanNuke dataset comprises 68,031 neoplastic and 23,207 epithelial cell images, while MoNuSAC comprises  33,104 lymphocytes, 1,252 neutrophils, and 1,695 macrophages, which we subsequently use in training cell classification models and classifying the cell image data \hyperref[fig:S2]{Appendix Figure S2 (1)}. 

The next step is to train two distinct ResNet50-based classifiers tailored to address the specific labeling challenges inherent in each dataset. We use ResNet50 for classification models due to its proven effectiveness for image classification tasks in histopathology \cite{pan2022reviewmachinelearningapproaches}, and its compatibility with small images. For the PanNuke dataset, we design the classifier, trained on MoNuSAC data, to disaggregate the heterogeneous "inflammatory" cell category into distinct subtypes: lymphocytes, macrophages, and neutrophils. Similarly, for the MoNuSAC dataset, the classifier is trained on PanNuke data and distinguishes between benign and malignant epithelial cells within the overarching "epithelial" label. By applying these targeted classifiers to their respective datasets, we assign more specific labels to individual cell instances, thus enabling us to create a unified dataset.
To ensure a balanced representation of classes, we train both models on datasets that had been equalized to match the size of the least represented class. Thus, we obtain datasets comprising 23,207 samples per class for PanNuke and 1,252 samples per class for MoNuSAC data. Next, we partition both of them into training (70\%), validation (20\%), and testing (10\%) subsets. To mitigate the risk of overfitting, we use a single dropout layer with a rate of p=0.5 in both models and data augmentation using randomized color perturbations, rotation, and horizontal and vertical flipping. We employ AdamW optimizer and the cross-entropy loss function for the training criterion.

To evaluate the two trained models, we measure the classification accuracy on the respective test subsets. The accuracies on the test subset for both classifiers are presented in \hyperref[tab:1]{Table 1}. The PanNuke model achieves an average accuracy of 93.57\%, with higher accuracy for neoplastic cells (96.06\%) compared to epithelial cells (86.26\%). The confusion matrix in Figure A3.1 shows that the model predominantly distinguishes accurately between epithelial and neoplastic tissues, with a substantial number of correct classifications and relatively few misclassifications. The MoNuSAC model demonstrates an average accuracy of 98.92\%, excelling in classifying lymphocytes (99.67\%) and macrophages (94.12\%), with lower performance for neutrophils (85.71\%). The confusion matrix in Figure A3.2 shows that the model identifies lymphocytes and performs reasonably well with macrophages and neutrophils.

\begin{table}[h!]
\renewcommand{\arraystretch}{1.5}
  \centering
  \caption{Cell classification results for PanNuke and MoNuSAC trained models (CI 95\%).}
  \label{tab:1}
  \begin{tabular}{|l|c|c|}
   \hline
    Accuracy               & PanNuke model              & MoNuSAC model              \\
    \hline
    Average      & 0.936 (0.931--0.941)         & 0.989 (0.986--0.993)        \\
    \hline
    Neoplastic   & 0.961 (0.956--0.965)         & -                          \\
    \hline
    Epithelial   & 0.863 (0.849--0.877)         & -                          \\
    \hline
    Lymphocytes  & -                          & 0.997 (0.995--0.999)        \\
    \hline
    Neutrophils  & -                          & 0.857 (0.796--0.918)        \\
    \hline
    Macrophages  & -                          & 0.941 (0.906--0.976)        \\
    \hline
  \end{tabular}
\end{table}

Finally, during the last step, we use the model trained on PanNuke data for epithelial cells in MoNuSAC and the model trained on MoNuSAC for the inflammatory cells class in PanNuke. Specifically, we use classifier models to relabel epithelial cells in MoNuSAC and inflammatory cells in PanNuke data. Then we combine cells with refined labels and the rest of the cells in both datasets to create a refined dataset (\hyperref[fig:S2]{Appendix Figure S2 (2)}). The process of relabeling cells and visualizing them on a patch is shown in \hyperref[fig:fig4]{Figure 4}. The cell counts in the refined dataset are provided in \hyperref[tab:S4]{Appendix Table S4}.

\begin{figure}[h!]
    \centering
    \includegraphics[width=\textwidth, height=0.42\textheight, keepaspectratio]{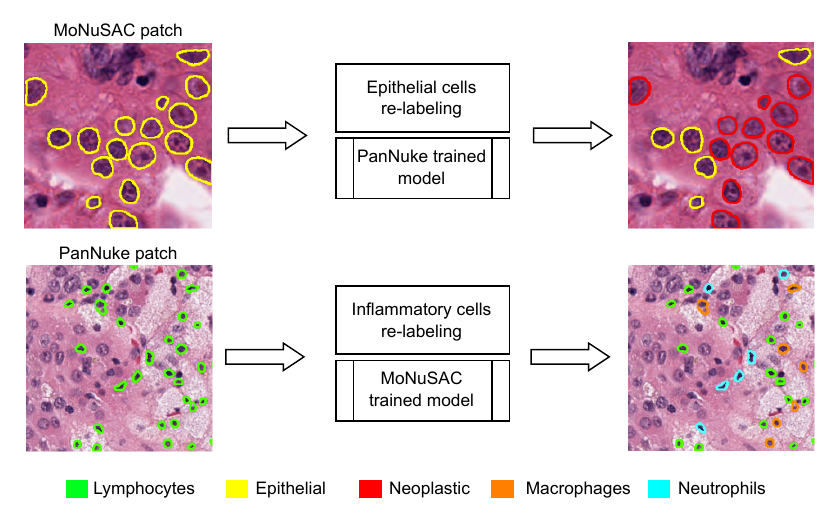}
    \caption{Cell relabeling procedure for epithelial and inflammatory cell classes}
    \label{fig:fig4}
\end{figure}


Relabeling and combining datasets have been explored in a prior study \cite{Parulekar_Kanwat_etal._2023}, where consecutive fine-tuning on multiple datasets was employed to account for hierarchical class label structures. While the method presented in \cite{Parulekar_Kanwat_etal._2023} is intuitive, it often lacks consistency and requires multiple fine-tuning runs, which can be cumbersome and time-consuming. 
In contrast, cross-relabeling simplifies this process by using specialized classification models tailored to each dataset's specific labeling challenges. This approach provides better transparency and produces a unified dataset encompassing seven distinct cell types across multiple tissue samples, enhancing data diversity for further model training or fine-tuning.

Despite these improvements, cross-relabeling does not entirely resolve issues related to poor labeling quality or the amount of labeled data. Specifically, our results show lower accuracies persist for underrepresented classes, such as macrophages, which may stem from a limited sample availability and intrinsic challenges in distinguishing these cells based solely on H\&E staining. Furthermore, while our method enhances label specificity, it relies on the initial quality of the broad labels; thus, any fundamental inaccuracies in the original annotations can propagate through the relabeling process. Addressing the overall problem of limited data labels may require integrating additional data sources or utilizing complementary immunohistochemical staining methods.
Although the reported performance metrics are obtained from evaluations on the native test sets of each dataset, it is important to note that the primary application of these classifiers is to perform cross-relabeling, where a model trained on one dataset (e.g., PanNuke) is applied to another (e.g., MoNuSAC) and vice versa. We acknowledge that a more systematic evaluation of cross-dataset generalization is needed and could be performed in future work.

Overall, the refined dataset produced by our approach can enhance the supervised training or fine-tuning of cell segmentation and classification models, especially those that utilize pre-trained foundation models to improve feature extraction robustness. In addition, these models can detect nuanced classes that enable researchers to conduct more detailed analyses of biological processes in computational pathology.

\section{Foundation models for robust cell segmentation and classification}

Accurate cell segmentation and classification in digital pathology are hindered by limited labeled data and the fact that conventional CNNs are unable to capture global contextual information due to their local receptive field constraints \cite{Gheflati_Rivaz_2022,Yang_Marcus_etal.}. Traditional approaches in cell quantification have predominantly relied on CNN encoders, such as ResNet50, given their proven effectiveness in semantic segmentation tasks \cite{Deshmane_2023,Graham_Vu_etal._2019,Mukasheva_Koishiyeva_etal._2024,Stringer_Wang_etal._2021}. However, approaches that include fine-tuning of pretrained CNNs, data augmentation, and stain normalization to partially increase data variability and address staining differences often fail to achieve the necessary generalization and robustness across diverse tissue types and staining conditions \cite{G._Wang_W._Li_etal._2018,Gao_Bagci_etal._2018,Karim_El_Khoury_Martin_Fockedey_etal._2021}.

\begin{figure}[h!]
    \centering
    \includegraphics[width=\textwidth]{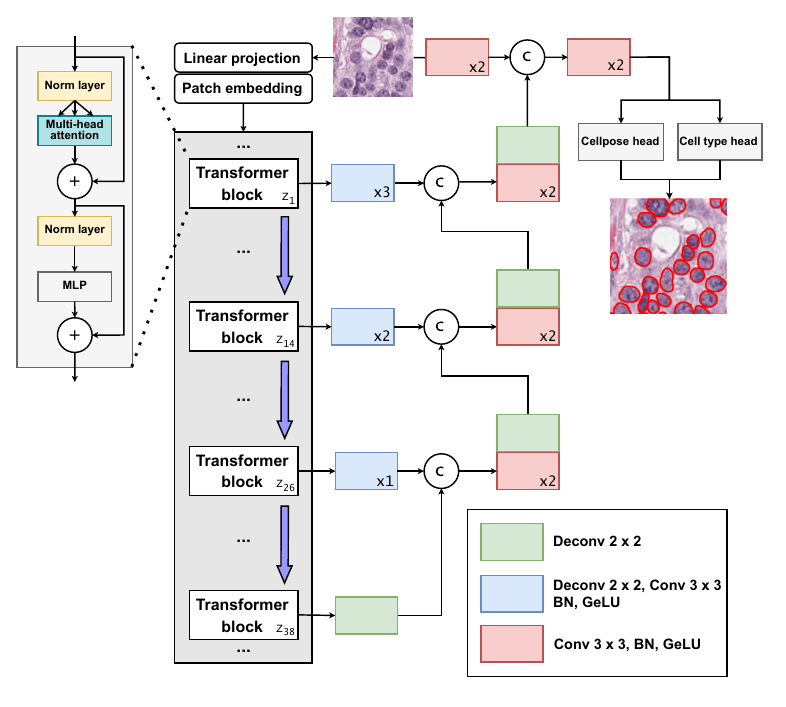}
    \caption{UNETR-like model with foundational model as backbone}
    \label{fig:fig5}
\end{figure}

To overcome these challenges, we leverage an encoder-decoder network that uses a foundation model as the encoder and a CNN upsampling decoder (\hyperref[fig:fig5]{Figure 5}) for simultaneous cell segmentation and classification in 2D patches extracted from WSIs. Foundation models with transformer-based architectures are viable alternatives to CNN-based encoders \cite{Shamshad_Khan_etal._2023,Sourget_2023}. They enable the creation of more advanced architectures that can decode or transform learned features more effectively \cite{Chen_Duan_etal._2023,Cheng_Misra_etal._2022,Xie_Wang_etal._2021}.

By utilizing a transformer-based encoder, we incorporate global contextual information into the feature extraction process, which is a key advantage of such architectures \cite{Chen_Lu_etal._2021}. This foundation model integration facilitates accurate pixel-wise segmentation and classification without the need for extensive encoder training, thereby potentially improving generalization across varied cellular structures and tissue types.
In our implementation, we employ a modified UNETR \cite{Hatamizadeh_Tang_etal._2021} architecture that combines a vision transformer (ViT) \cite{Dosovitskiy_Beyer_etal._2021} encoder with a CNN-based decoder. The encoder utilizes the pretrained H-Optimus foundation model, which contains 1.1 billion parameters and is trained on over 500,000 H\&E stained WSIs \cite{Saillard_Jenatton_etal._2024}. We extract outputs from four evenly spaced transformer blocks $Z_i$, where $i \in [1, 14, 26, 38]$, to serve as residual connections for the CNN decoder. We select these blocks based on our observation that features from non-adjacent levels of the encoder lead to better overall performance on the test subset.

The CNN decoder upsamples the feature representations, acquired from the transformer blocks, to generate an intermediate vector that is handled by two task-specific layers that generate cell segmentation and classification masks. The first task-specific layer is the ‘Cellpose head’,  which is used to delineate cell instances. The layer generates horizontal and vertical gradient maps to form vector fields that are refined through gradient tracking in a post-processing step using the Cellpose algorithm \cite{Stringer_Wang_etal._2021}, known for its efficacy in cell segmentation tasks and generalizability across multiple domains \cite{Pachitariu_Stringer_2022,Stringer_Pachitariu_2024}. The second task-specific layer is the "Cell type head", which assigns labels to individual pixels. In the post-processing step, we determine the output classification label of each segmented cell instance by majority voting over the labeled pixels that comprise the cell in the segmentation map.

To evaluate model performance and measure the impact of adding a foundation model as backbone, we compare it to a ResNet50-based model. ResNet50 is a widely used solution for encoders in segmentation architectures in the medical domain \cite{Deshmane_2023,Graham_Vu_etal._2019,Mukasheva_Koishiyeva_etal._2024,Stringer_Wang_etal._2021}. For the H-Optimus-based model, we utilize frozen weights for the encoder and only fine-tune the decoder to take advantage of the extensive pre-training of the foundation model. For the ResNet50-based model we start with ImageNet \cite{Deng_Dong_etal.} weights and train both encoder and decoder parts. Hyperparameters for the training step are set to be identical, where possible, for comparable evaluation. 
For this evaluation, we deliberately use the PanNuke dataset to provide a standardized and controlled comparison between the H‑Optimus and ResNet50-based models (\hyperref[fig:S2]{Appendix Figure S2 (3)}). Specifically, we use two of the default PanNuke dataset splits (66\%) for training and validation, and reserve the third split (33\%) for testing.

To address the challenge of cell class imbalance in the PanNuke dataset, which is a common characteristic in most cell-level H\&E patch datasets, both models’ training processes employ a weighted loss function comprising cross-entropy and focal loss \cite{Lin_Goyal_etal._2018}. The focal loss component is adjusted with coefficients derived from each cell class' instance frequency, emphasizing learning from underrepresented classes and enhancing the model's sensitivity to rare but significant cellular patterns. The cross-entropy loss is augmented with spectral decoupling regularization \cite{Pezeshki_Kaba_etal._2021,Pohjonen_Stürenberg_etal._2022} and spatially varying label smoothing \cite{Islam_Glocker_2021}, which potentially stabilizes training and improves generalization in case of complex tissue morphologies. For optimization, we employ the \textit{AdamW} \cite{Loshchilov_Hutter_2019} to counter unbalanced class scenarios, with cosine annealing learning rate scheduler.

We utilize the scikit-learn library \cite{Van_der_Walt_Schönberger_etal._2014} and HoVer-Net \cite{Graham_Vu_etal._2019} implementations of $R^2$ (the coefficient of determination) and $PQ$ (panoptic quality) to evaluate our experiments. Complete mathematical formulations and detailed explanations of these metrics are provided in \hyperref[chap:S5]{Appendix S5}. To compute confidence intervals, we use nonparametric bootstrapping, where after calculating the metric on the full sample, we generated 1000 bootstrap replicates by resampling with replacement and then determined the 95\% confidence intervals as the 2.5th and 97.5th percentiles of the resulting empirical distribution.


The model comparisons are summarized in \hyperref[tab:2]{Table 2}. The H‑Optimus-based model achieves higher $R^2$ across all cell classes compared to the ResNet50-based model, which means that its predictions are more closely aligned with the PanNuke cell counts, indicating a stronger correlation with the observed data. Notably, the improvement of $R^2_{dead}$ may be an indicator of better global contextual representations provided by the foundation model backbone. In terms of segmentation and classification quality combined, measured by the PQ score, the H‑Optimus-based model demonstrates notable improvements across most cell classes. Overall, the average $R^2$ improved from 0.575 to 0.871, while the average $PQ$ score improved from 0.450 to 0.492, demonstrating better performance of the H-Optimus-based model.

\begin{table}[h!]
\renewcommand{\arraystretch}{1.5}
  \centering
  \caption{Cell quantification metrics for baseline and proposed models (CI 95\%).}
  \label{tab:2}
  \begin{tabular}{|l|c|c|}
    \hline
    Metric             & Resnet50-based            & H-optimus-based              \\
    \hline
    $R^2_{neoplastic}$    & 0.681 (0.576--0.769)       & \textbf{0.941 (0.917--0.960)} \\
    \hline
    $R^2_{inflammatory}$  & 0.863 (0.778--0.903)       & \textbf{0.949 (0.918--0.966)} \\
    \hline
    $R^2_{connective}$    & 0.600 (0.488--0.698)       & 0.609 (0.436--0.772)          \\
    \hline
    $R^2_{dead}$          & 0.097 (-11.389--0.669)     & 0.925 (0.404--0.982)          \\
    \hline
    $R^2_{epithelial}$    & 0.635 (0.490--0.747)       & \textbf{0.930 (0.886--0.964)} \\
    \hline
    $PQ_{neoplastic}$       & 0.517 (0.499--0.535)       & \textbf{0.589 (0.575--0.604)} \\
    \hline
    $PQ_{inflammatory}$     & 0.455 (0.429--0.482)       & \textbf{0.528 (0.507--0.549)} \\
    \hline
    $PQ_{connective}$       & 0.416 (0.400--0.431)       & \textbf{0.451 (0.436--0.465)} \\
    \hline
    $PQ_{dead}$             & 0.374 (0.342--0.408)       & 0.292 (0.209--0.365)          \\
    \hline
    $PQ_{epithelial}$       & 0.488 (0.460--0.519)       & \textbf{0.599 (0.579--0.618)} \\
    \hline
  \end{tabular}
\end{table}

Our results  show that integrating the H‑Optimus foundation model within the UNETR architecture enhances the model's ability to segment and classify cells across diverse tissues from PanNuke data. The pretrained transformer encoder provides robust feature representations, resulting in higher average $R^2$ and $PQ$ scores compared to the CNN-based model. This leads to more reliable cell quantification and more accurate downstream analysis. Additionally, the streamlined fine-tuning process reduces computational overhead and training time, making the model more adaptable for new data.

Despite these advancements, the foundation model-based approach does not fully resolve all challenges related to cell segmentation and classification. We observe lower metric scores for underrepresented classes in the training data. Furthermore, foundation models typically encompass billions of parameters, resulting in substantial computational and memory requirements. It therefore poses challenges for deployment in resource-constrained environments, limiting their practical applicability in certain clinical settings.

\section{Model optimization via Knowledge Distillation}

\begin{figure}[h!]
    \centering
    \includegraphics[width=\textwidth, height=0.45\textheight, keepaspectratio]{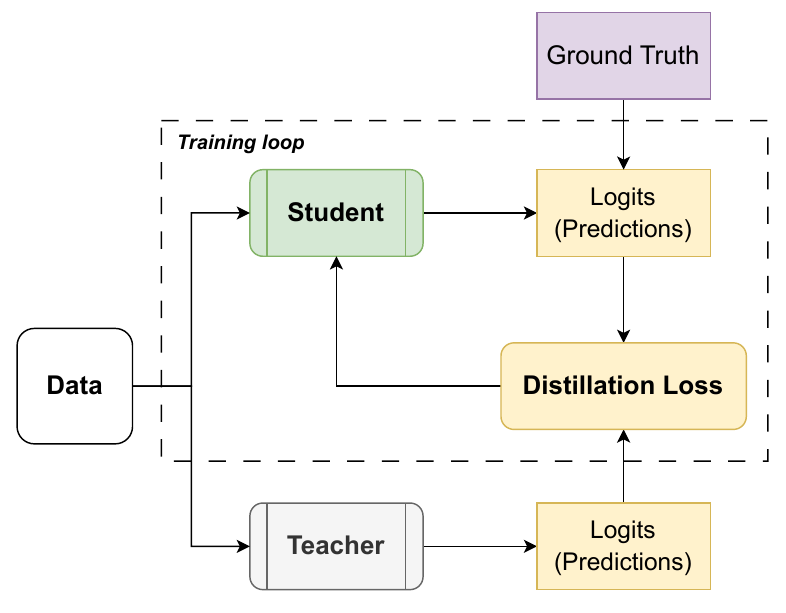}
    \caption{Knowledge distillation framework for training a student model using a pre-trained teacher}
    \label{fig:fig6}
\end{figure}

To address the limitations posed by the extensive size of foundation models, we implement knowledge distillation — a model compression technique that leverages the teacher-student paradigm \cite{Hinton_Vinyals_etal._2015}. By training a smaller, more efficient student model to replicate the output of a larger, pre-trained teacher model, we retain performance while significantly reducing the model's complexity and resource requirements (\hyperref[fig:fig6]{Figure 6}).

We employ knowledge distillation to compress the H‑Optimus-based teacher model into a more efficient student model. The teacher model is the modified UNETR architecture with the H‑Optimus foundation model described in the previous chapter. The student model is based on a UNet architecture augmented with residual connections and incorporates a smaller ViT encoder with 9 million parameters \cite{Steiner_Kolesnikov_etal._2022,Wightman_2019}. 

First, we fine-tune the teacher model using the refined dataset from the cross-relabeling procedure (Section 2). Initially we train the decoder of the teacher model while keeping the encoder weights frozen. We split the refined dataset into train (70\%), validation (20\%) and test (10\%) subsets (\hyperref[fig:S2]{Appendix Figure S2 (4)}). During fine-tuning, we use the train and validation subsets, while leaving the test subset for model evaluation. We set the training procedure and model hyperparameters to be identical to those that were used to demonstrate the utility of foundation models for the simultaneous cell segmentation and classification task.

Next, we perform knowledge distillation from teacher to student using the refined dataset used to fine-tune the teacher model. The student model is trained to replicate the teacher model's outputs. We utilize a specialized loss function that aligns the student's predicted probability distribution with the teacher's, incorporating the teacher's class probability distribution derived from the output. Following the methodology of Hinton et al. \cite{Hinton_Vinyals_etal._2015}, we experiment with various hyperparameter settings for the temperature ($T$) and the balancing coefficients ($\alpha$ and $\beta$) in the loss function. We vary $T$ from 1 to 20 and adjust $\alpha$ and $\beta$ to balance the distillation and student losses. Through iterative tuning and evaluation, we identify that setting $T=14$, $\alpha=0.3$, and $\beta=0.7$ yields a configuration that converges and closely approximates the teacher model's performance during training.

Finally, we assess the performance of both models using the $R^2$ and $PQ$ (defined in \hyperref[chap:S5]{Appendix S5}) on the test set of the refined dataset (\hyperref[tab:3]{Table 3}). We observe that the 95\% confidence intervals overlap for most cell types, so we cannot claim statistically significant performance differences between the teacher and student models. One exception appears in the neoplastic class. The teacher model produces an $R^2$ of 0.919, while the student model shows an $R^2$ of 0.852. In addition, the student model achieves higher $PQ$ values for the neoplastic and connective classes, though the confidence intervals show overlap.

\begin{table}[h!]
\renewcommand{\arraystretch}{1.5}
  \centering
  \caption{Cell quantification metrics for teacher and distilled student models (CI 95\%).}
  \label{tab:3}
  \begin{tabular}{|l|c|c|}
    \hline
    Metric & Teacher & Student \\
    \hline
    $R^2_{neoplastic}$    & \textbf{0.919} (0.898--0.939) & 0.852 (0.800--0.891) \\
    \hline
    $R^2_{lymphocyte}$    & 0.969 (0.956--0.977)         & 0.969 (0.956--0.978) \\
    \hline
    $R^2_{connective}$    & 0.694 (0.548--0.809)         & 0.618 (0.469--0.741) \\
    \hline
    $R^2_{dead}$          & 0.755 (0.400--0.908)         & 0.424 (0.100--0.731) \\
    \hline
    $R^2_{epithelial}$    & 0.922 (0.870--0.958)         & 0.843 (0.738--0.917) \\
    \hline
    $R^2_{macrophage}$    & 0.384 (-0.369--0.724)        & 0.704 (0.352--0.859) \\
    \hline
    $R^2_{neutrofil}$     & 0.854 (0.578--0.929)         & 0.833 (0.502--0.925) \\
    \hline
    $PQ_{neoplastic}$       & 0.581 (0.569--0.593)         & 0.601 (0.588--0.613) \\
    \hline
    $PQ_{lymphocyte}$       & 0.536 (0.520--0.553)         & 0.563 (0.544--0.579) \\
    \hline
    $PQ_{connective}$       & 0.436 (0.421--0.451)         & 0.457 (0.441--0.474) \\
    \hline
    $PQ_{dead}$             & 0.272 (0.235--0.315)         & 0.279 (0.201--0.369) \\
    \hline
    $PQ_{epithelial}$       & 0.522 (0.500--0.545)         & 0.530 (0.506--0.555) \\
    \hline
    $PQ_{macrophage}$       & 0.524 (0.459--0.588)         & 0.474 (0.405--0.543) \\
    \hline
    $PQ_{neutrofil}$        & 0.541 (0.490--0.592)         & 0.565 (0.522--0.607) \\
    \hline
  \end{tabular}
\end{table}

We further decompose the $PQ$ metric into its $SQ$ and $DQ$ components (\hyperref[tab:S6]{Appendix Table S6}). Both models produce nearly identical $SQ$ values, which indicates that they predict instance boundaries with similar precision. Although the student model shows some improvement in $DQ$ scores for certain classes, the confidence intervals overlap and do not confirm a statistically significant difference.

We observe that the student and teacher models yield comparable detection performance despite the student model using a much smaller and simpler architecture. A model with fewer parameters reduces the risk of overfitting when training data are scarce relative to the model’s complexity \cite{Farias_Ludermir_etal._2022}. The knowledge distillation process also encourages the student model to focus on the most generalizable detection features learned from the teacher. These factors enable the student model to achieve similar detection performance across different cell types.

Additionally, considering the model sizes reported in \hyperref[tab:4]{Table 4}, the distilled model achieves a significant reduction compared to the teacher model, with a 48-fold decrease in parameter count and a 5.5-fold reduction in on-disk size. In inference mode, the teacher model requires 16 GB of VRAM for a batch size of 32, while the distilled model only needs 3 GB of VRAM for the same batch size. These reductions make the distilled model significantly more practical for fine-tuning and deployment in resource-constrained environments.

\begin{table}[h!]
\renewcommand{\arraystretch}{1.5}
  \centering
  \caption{Parameter counts and size of teacher and distilled model}
  \label{tab:4}
  \adjustbox{max width=\textwidth}{%
  \begin{tabular}{|l|c|c|c|}
    \hline
    Metric & H-optimus-based (Teacher) & mobileViT-based (Student) & Magnitude of difference \\
    \hline
    Parameters count       & 1,158,917,906   & \textbf{24,093,393}   & \textbf{48x}  \\
    \hline
    Estimated Total Size (MB) & 87,912       & \textbf{15,935}    & \textbf{5.5x} \\
    \hline
  \end{tabular}%
}
\end{table}


With recent advancements in complex network architectures and the use of pretrained encoders to achieve state-of-the-art performance \cite{Baumann_Dislich_etal._2024,Hörst_Rempe_etal._2024} in cell segmentation and classification tasks, model size, computational complexity, and processing times have increased. This limits the scalability and accessibility of these models. As we demonstrate, this may be mitigated using knowledge distillation. Studies in the field of natural language processing have demonstrated the efficacy of knowledge distillation in retaining the capabilities of the teacher model while achieving significant reductions in size and complexity \cite{Huangpu_Gao_2024,Sun_Yu_etal.}. 

We demonstrate the feasibility of knowledge distillation in digital pathology, specifically for cell segmentation and classification tasks. Moreover, we achieve this performance while also significantly reducing the parameter count. In addressing the challenge of knowledge transfer, we found that distillation from a transformer-based model to a smaller transformer is more straightforward than attempting to map transformer features to CNN blocks. In our experiments, using a CNN-based network as a student results in worse cell quantification performance due to the structural constraints of CNN feature space dimensions. 

Although our primary approach relies on a transformer-based student model that performs well, it can be further optimized to incorporate advantages from CNN architectures. For example, employing alternative techniques such as using ViT adapters \cite{Chen_Duan_etal._2023} or $1 \times 1$ convolutions to adjust feature map sizes may be beneficial for harnessing CNN advantages like enhanced local feature extraction. Moreover, if additional performance improvements are desired, the process can be further enhanced by applying supplementary knowledge distillation techniques, such as self-distillation \cite{Zhang_Song_etal._2019} or online distillation \cite{Houyon_Cioppa_etal._2023}.

Despite these promising results, further validation on independent datasets is necessary to fully understand the model's limitations. Underrepresented classes may pose challenges when addressing complex cases. Pathologists need to validate these models to adopt them in clinical settings. While the distilled models are smaller and more deployable, a technological gap persists because pathologists traditionally rely on established methods for inspecting WSIs and diagnosing diseases. Addressing the complexities involved in deploying models for inference and supporting pathologists in adopting new tools is essential for integrating these models into clinical workflows.

\section{Model integration with QuPath}
\begin{figure}[h!]
    \centering
    \includegraphics[width=\textwidth]{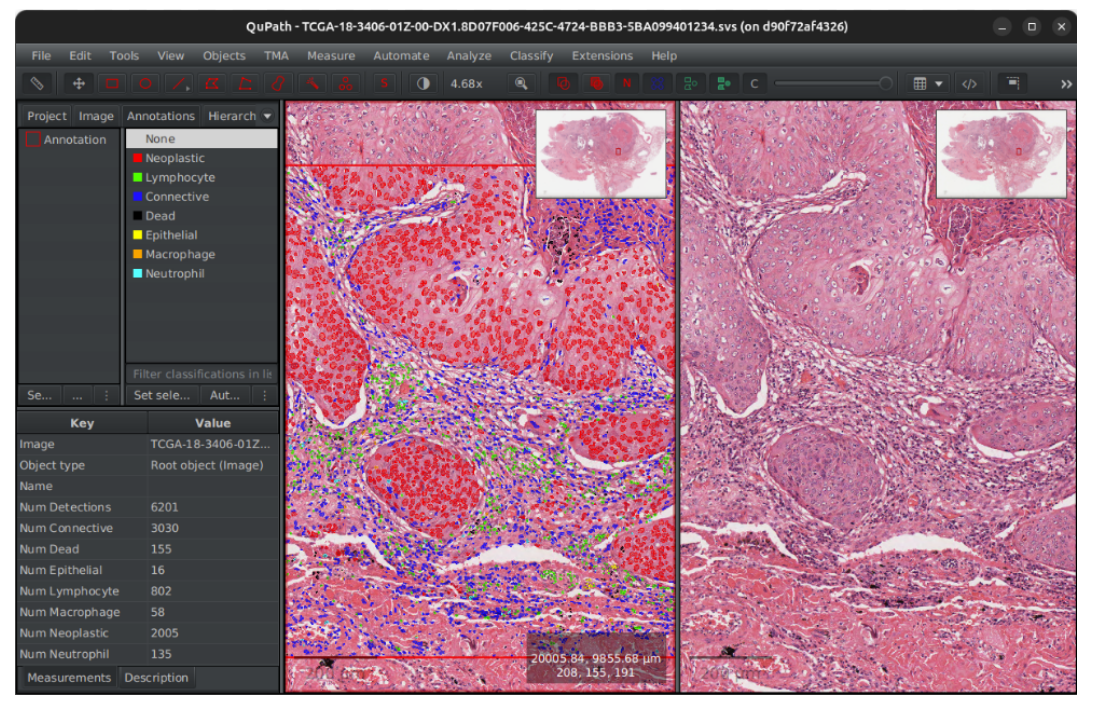}
    \caption{Visualization of model-generated cell quantification annotations (left) and the corresponding unannotated slide (right) in QuPath}
    \label{fig:fig7}
\end{figure}

Digital pathology tools with graphical user interfaces are essential for visualizing and analyzing WSIs. To make our student model useful in clinical pathology workflows, it needs to be integrated into a tool that enables inspecting regions, creating annotations, and providing quantitative analyses of biomarkers. Therefore, we integrate the trained student model from the previous chapter into the QuPath open‑source platform \cite{Bankhead_Loughrey_etal._2017}. QuPath provides the required annotation, visualization, and analysis tools to interpret complex histological data, including workflows for cell segmentation, classification, and quantification (\hyperref[fig:fig7]{Figure 7}). 

To identify the regions in a WSI critical for prognosticating tumor development, such as specific tumor areas or border regions without overlapping healthy tissue, the pathologist uses QuPath to outline these regions. Then, the pathologist initiates a cell segmentation and classification script through the QuPath interface for the selected regions. The resulting annotations and quantified cell information are then directly overlaid onto the WSI in the QuPath interface. Additional design and implementation details are in \hyperref[chap:S7]{Appendix S7}. 

Two common approaches for integrating deep learning models into QuPath are Java‑based native QuPath extensions \cite{Goldsborough_Philps_etal._2024} and the execution of RESTful API requests to a model server coupled with handling the response via an extension, as demonstrated in the application of cell segmentation models applied to immunofluorescence images \cite{Sugawara_2023}. While the community is actively working on these integration strategies, there is currently no universal solution that fully addresses all integration and performance requirements.

Extensions may offer better integration with QuPath, allowing slightly improved performance and more widespread usage of the built-in QuPath models, but they lack the flexibility to customize models and modify their behavior. For example, the newest version of QuPath includes models such as StarDist \cite{Weigert_Schmidt} and InstanSeg \cite{Goldsborough_Philps_etal._2024} that can perform cell segmentation. Both models pose limitations when applied to simultaneous cell segmentation and classification. StarDist performs well only on convex, round shapes by design, whereas some neoplastic, inflammatory, and connective cells exhibit complex and non-convex shapes. InstanSeg provides only semantic segmentation without assigning classes to the segmented cells.


In contrast, our approach offers an alternative integration strategy. It utilizes the paquo library to directly interact with QuPath’s internal application programming interface from within Python. This enables data exchange and processing without the need for intermediate conversion steps and provides greater control over model customization, retraining, and the incorporation of custom processing steps.

The integration of our custom model with QuPath underscores its potential to significantly enhance the diagnostic process by reducing the time burden on pathologists and enabling them to focus on more complex interpretative tasks using familiar software. Leveraging a tool that is already well-established among pathologists increases the likelihood of its adoption into daily clinical workflows. The quantitative data generated through the automated workflow is critical for both clinical decision-making and research, facilitating more accurate biomarker analysis, enabling robust statistical evaluations, and supporting hypothesis generation and testing. Additionally, by streamlining cell segmentation and classification, the tool enhances the scalability and reproducibility of pathological assessments, ultimately contributing to improved diagnostic accuracy and patient outcomes.

\section{Conclusion and future work}

In this study, we address critical challenges in digital pathology and tackle the usability and deployment issues of the developed models in standard computing environments without the need for high-performance computing systems. Our multi-faceted approach encompasses data refinement through cross-relabeling, leveraging foundation models for robust cell segmentation and classification, optimizing model performance via knowledge distillation, and integrating the optimized model into the QuPath software for practical application. This approach is used to construct a capable, versatile, and adjustable model for cell segmentation and classification, with enhanced performance and usability.

\begin{sloppypar}
While our approach shows potential in the field of computational pathology, certain limitations persist. 
For example, our implementation currently exhibits lower performance in detecting macrophages. 
This serves as an instance of the broader challenge of accurately identifying complex cell types. In order to address this issue, extending our approach to incorporate additional data sources, exploring alternative modeling approaches, and integrating other imaging modalities such as immunohistochemical staining may help improve detection accuracy. Moreover, although the distilled model reduces computational demands, integrating advanced deep learning models into clinical practice requires addressing technological gaps and potential resistance to adopting new tools within established diagnostic processes.
\end{sloppypar}

Future work could focus on several key areas to refine the proposed approach and facilitate its adoption in clinical environments. Enhancing the cell-relabeling process with additional datasets \cite{Graham_Jahanifar_etal._2021} could improve the representation of underrepresented cell types and enhance overall model performance. Also, incorporating additional data sources, such as multi-modal imaging or complementary staining methods, may address limitations related to cell type differentiation and class imbalance. Exploring other foundation models \cite{Vorontsov_Bozkurt_etal._2024,Zimmermann_Vorontsov_etal._2024} or introducing additional modalities \cite{Ding_Wagner_etal._2024,Vaidya_Zhang_etal._2025} may provide alternative architectures better suited to specific tasks or offer improved efficiency. Implementing more complex knowledge distillation techniques \cite{Houyon_Cioppa_etal._2023,Zhang_Song_etal._2019} could further optimize the model's performance and adaptability. Additionally, deeper integration with QuPath or other digital pathology software could provide pathologists more control over cell quantification analysis directly within the QuPath interface, thereby increasing accessibility and usability. Such enhancements would not only refine model performance but also ensure greater adaptability and scalability within various clinical environments. Finally, extensive validation of the model by pathologists and benchmarking against independent datasets are essential steps toward establishing the model's reliability and fostering confidence in its clinical utility.

\section*{Acknowledgments} 
This work was funded in part by the Research Council of Norway grant no. 309439 SFI Visual Intelligence, and the North Norwegian Health Authority grant no. HNF1521-20.

\bibliographystyle{IEEEtran}
\begin{sloppypar}

\end{sloppypar}

\clearpage
\beginsupplement
\section*{Appendix}
\renewcommand{\thesubsection}{S\arabic{subsection}}

\subsection{\label{chap:S1}PanNuke and MoNuSAC preprocessing}
The PanNuke dataset comprises a set of 7,901 RGB patches, each with dimensions of $256 \times 256$ pixels, which we set as the standard patch size for our analysis. In contrast, the MoNuSAC dataset encompasses 294 images of heterogeneous dimensions. To standardize the MoNuSAC images with our experiments, we implement a standardization protocol. Specifically, for images exceeding the dimensions of $256 \times 256$ pixels, we segment them into equal-sized patches and apply mirror padding to the remaining portions to avoid information loss at the peripherals. Patches with dimensions less than $128 \times 128$ pixels are excluded from the dataset due to the insufficient resolution to capture relevant cellular details. For patches where either dimension falls between 128 and 256 pixels, we employ upsampling to achieve the standard patch size. As a result, we obtain a total of 2,823 RGB patches derived from the MoNuSAC dataset for subsequent analysis. For additional details on the MoNuSAC data preparation process, refer to the source code \cite{Shvetsov_2025a}.
\clearpage

\subsection{\label{chap:S2}Data usage for the methodology}

\counterwithin{figure}{subsection}
\renewcommand{\thefigure}{S\arabic{subsection}}

\begin{figure}[h!]
    \centering
    \includegraphics[width=\textwidth, height=0.85\textheight, keepaspectratio]{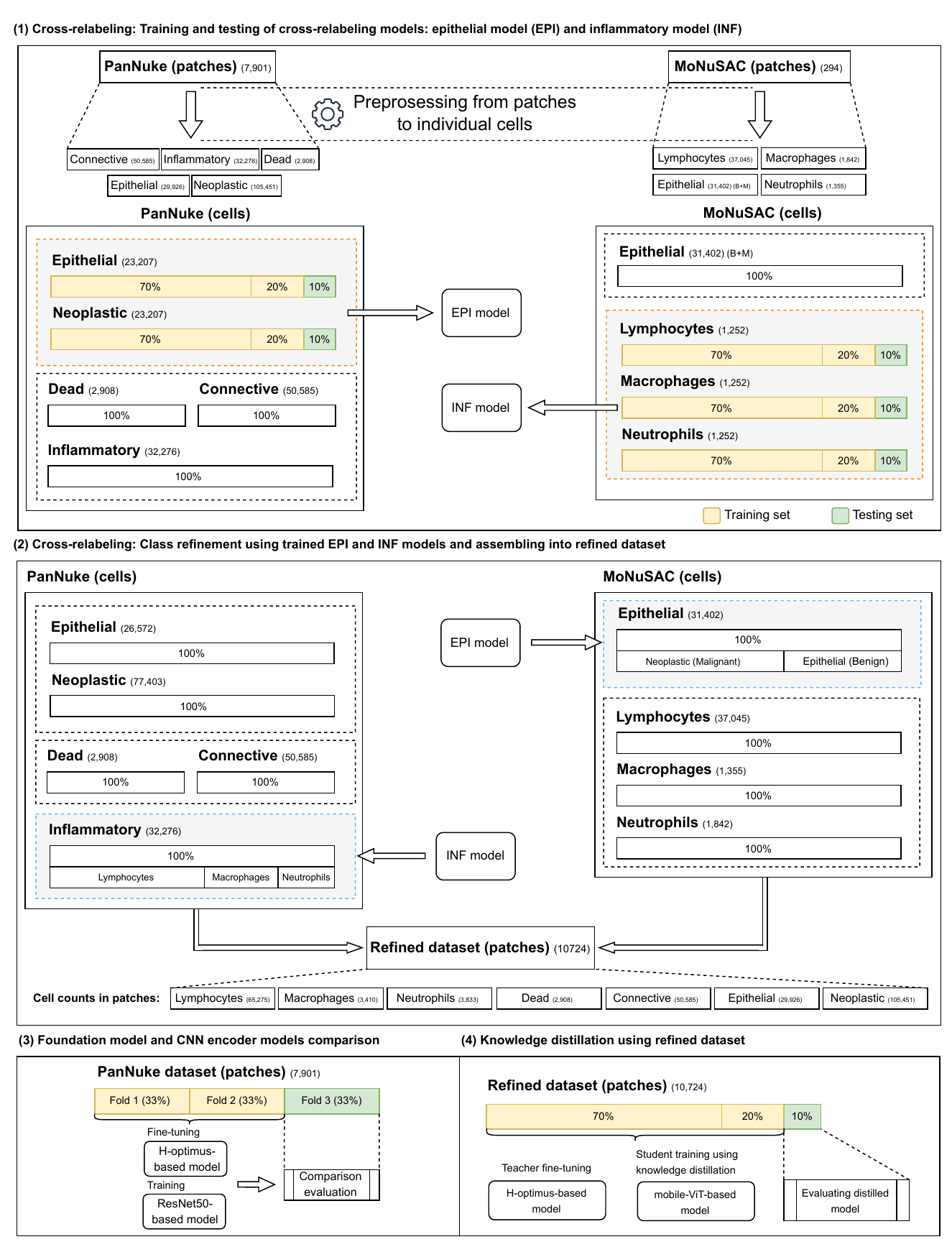}
    \caption{Overview of the methodology for cross-labeling, dataset refinement, and model comparison. (1) Cross-relabeling - training and testing cell classification models, (2) Cross-relabeling - using cell classification models to create refined dataset, (3) Fine-tuning and training models for comparison, (4) Student knowledge distillation with refined dataset}
    \label{fig:S2}
\end{figure}
\clearpage

\subsection{\label{chap:S3}Confusion matrices for classification models}
\counterwithin{figure}{subsection}
\renewcommand{\thefigure}{S\arabic{subsection}.\arabic{figure}}

\begin{figure}[h!]
    \centering
    \includegraphics[width=\textwidth, height=0.4\textheight, keepaspectratio]{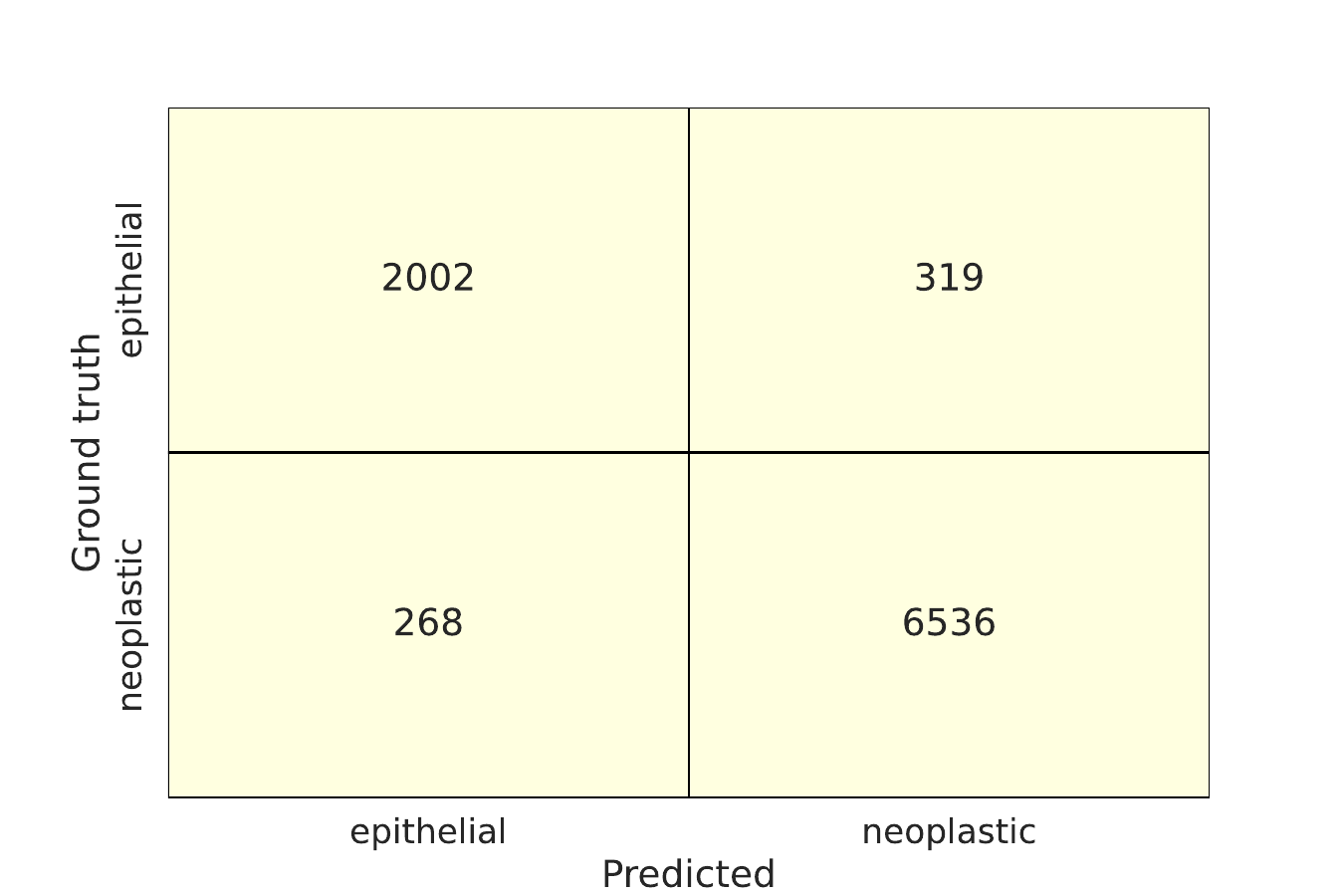}
    \caption{Confusion matrix for PanNuke trained model}
    \label{fig:S3.1}
\end{figure}

\begin{figure}[h!]
    \centering
    \includegraphics[width=\textwidth, height=0.4\textheight, keepaspectratio]{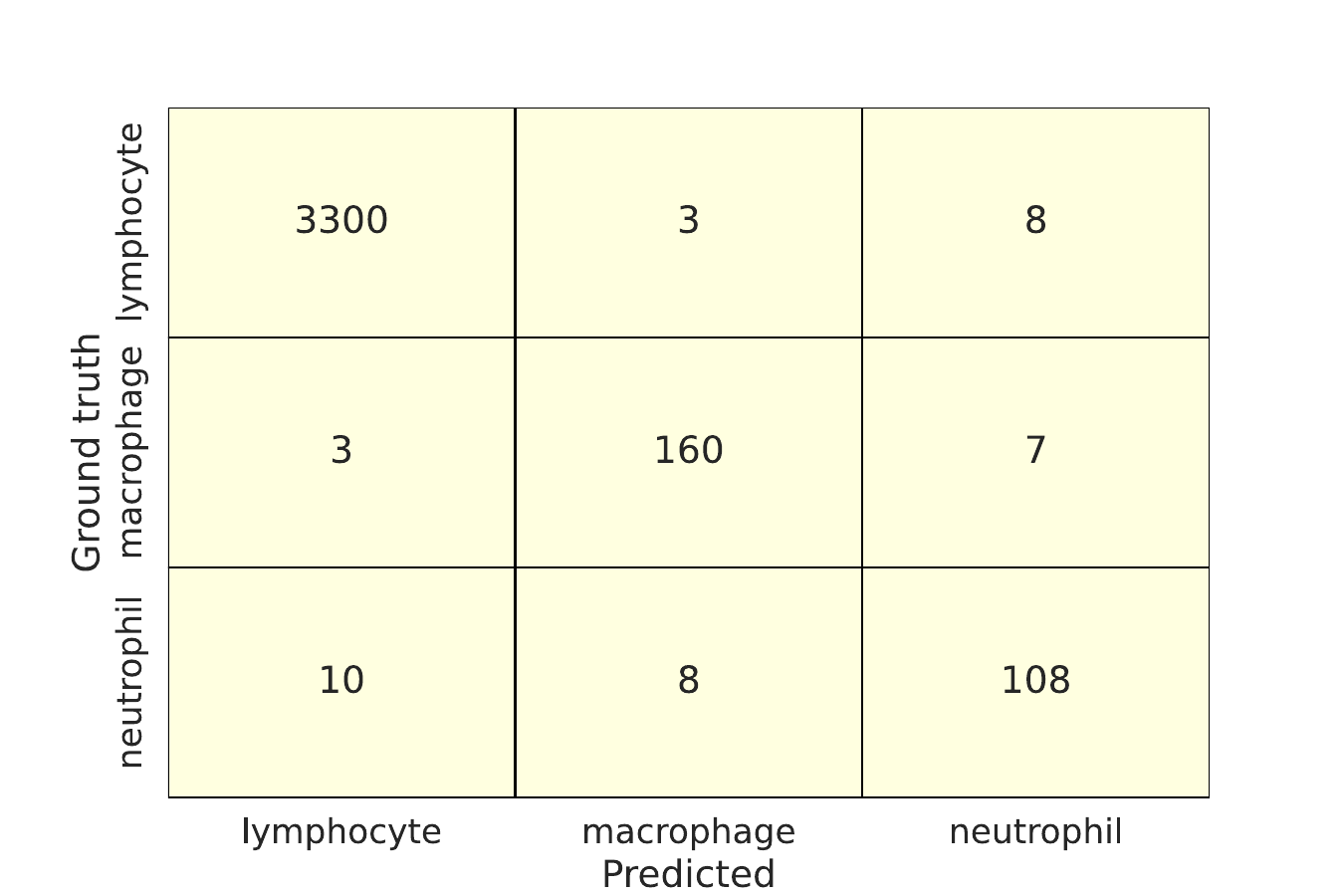}
    \caption{Confusion matrix for MoNuSAC trained model}
    \label{fig:S3.2}
\end{figure}

\clearpage

\subsection{\label{chap:S4}Datasets cell counts}

\counterwithin{table}{subsection}
\renewcommand{\thetable}{S\arabic{subsection}}

\begin{table}[h!]
\renewcommand{\arraystretch}{2.0}
\centering
\caption{\label{tab:S4}Cell counts for PanNuke, MoNuSAC and refined datasets. Numbers in parentheses indicate preprocessed cell counts for cell classifier models training and testing.}
\begin{tabular}{|l|c|c|c|}
\hline
Cell type & PanNuke & MoNuSAC & Refined \\
\hline
Neoplastic & 77,403 (68,031) & - & 105,451 \\
\hline
Epithelial & 26,572 (23,207) & - & 29,926 \\
\hline
Epithelial (benign and malignant) & - & 31,402 & - \\
\hline
Inflammatory & 32,276 & - & - \\
\hline
Lymphocytes & - & 37,045 (33,104) & 65,275 \\
\hline
Neutrophils & - & 1,355 (1,252) & 3,833 \\
\hline
Macrophage & - & 1,842 (1,695) & 3,410 \\
\hline
Dead & 2,908 & - & 2,908 \\
\hline
Connective & 50,585 & - & 50,585 \\
\hline
\end{tabular}
%
\end{table}

\clearpage

\subsection{\label{chap:S5}Definition of validation metrics}
\counterwithin{equation}{subsection}
\renewcommand{\theequation}{\arabic{equation}}

\subsubsection{\label{chap:S5.1}R\textsuperscript{2}}
The coefficient of determination, denoted as $R^2$, is a statistical measure that represents the proportion of variance in the dependent variable that is predictable from the independent variables. In the context of cell quantification in pathology, $R^2$ is used to assess how well the predicted quantities of different cell types in a patch align with the actual quantities observed in the ground truth data, with higher values representing more accurate quantification. $R^2$ is defined as
\begin{equation*}
R^2 = 1 - \frac{\sum_{i=1}^n (y_i - \hat{y}_i)^2}{\sum_{i=1}^n (y_i - \bar{y})^2},
\end{equation*}
where $y_i$ represents the actual number of cells of a specific type in the $i$-th image, $\hat{y}_i$ represents the predicted number of cells of that type in the $i$-th image, $\bar{y}$ is the mean of the actual numbers across all images, and $n$ is the total number of images in the dataset.

The $R^2$ metric has a range of $(-\infty, 1]$. An $R^2$ of 1 indicates perfect prediction, where all predicted values exactly match the actual values. An $R^2$ of 0 suggests that the model explains none of the variability of the response data around its mean. If $R^2$ is negative, it indicates that the model performs worse than a model that simply predicts the mean of the actual values for all observations.

\subsubsection{\label{chap:S5.2}PQ}
Panoptic Quality ($PQ$) is a comprehensive metric used to evaluate the performance of segmentation models in tasks that require both instance segmentation and classification. $PQ$ provides a single score that encapsulates both the detection accuracy (i.e., how many objects were correctly identified) and the segmentation quality (i.e., how accurately the objects' boundaries were delineated). This metric is particularly useful in multiclass scenarios where each pixel is classified into distinct categories, such as different cell types in pathology images.

$PQ$ is calculated as the product of two terms: Detection Quality ($DQ$) and Segmentation Quality ($SQ$). It can be expressed as
\begin{equation*}
PQ = DQ \cdot SQ,
\end{equation*}
where
\begin{equation*}
DQ = \frac{TP}{TP + 0.5\, FP + 0.5\, FN},
\end{equation*}
\begin{equation*}
SQ = \frac{\sum_{(p, g) \in \mathcal{M}} IoU(p, g)}{TP}.
\end{equation*}
In these formulas, $TP$ denotes the number of correctly matched instances between ground truth and prediction, $FP$ denotes the predicted instances that have no corresponding ground truth, $FN$ denotes the ground truth instances that were not detected, $IoU(p, g)$ is the Intersection over Union for a pair of matched instances $p$ (prediction) and $g$ (ground truth), and $\mathcal{M}$ is the set of matched pairs.

The $PQ$ metric is calculated for each class and is averaged across classes to provide a global performance measure.

The $PQ$ score has a range of $[0, 1.0]$, where a higher score indicates better performance in both detecting and segmenting the instances correctly. A $PQ$ of 1 signifies perfect identification and segmentation of all instances, whereas a $PQ$ of 0 indicates that no instances were correctly identified and segmented.

\clearpage

\subsection{\label{chap:S6}Segmentation and Detection quality metrics for teacher and student models}

\begin{table}[h!]
\renewcommand{\arraystretch}{2.0}
\centering
\caption{Segmentation and detection quality for student and teacher models (CI 95\%)}
\label{tab:S6}
\begin{tabular}{|l|c|c|}
\hline
Metric & Teacher & Student \\
\hline
$SQ_{neoplastic}$ & 0.819 (0.815--0.823) & 0.824 (0.819--0.828) \\
\hline
$SQ_{lymphocyte}$ & 0.795 (0.788--0.802) & 0.790 (0.783--0.796) \\
\hline
$SQ_{connective}$ & 0.770 (0.762--0.776) & 0.780 (0.772--0.786) \\
\hline
$SQ_{dead}$ & 0.659 (0.623--0.688) & 0.657 (0.624--0.695) \\
\hline
$SQ_{epithelial}$ & 0.780 (0.770--0.790) & 0.788 (0.779--0.797) \\
\hline
$SQ_{macrophage}$ & 0.788 (0.760--0.810) & 0.757 (0.730--0.783) \\
\hline
$SQ_{neutrofil}$ & 0.782 (0.761--0.801) & 0.775 (0.759--0.792) \\
\hline
$DQ_{neoplastic}$ & 0.706 (0.692--0.719) & 0.727 (0.712--0.741) \\
\hline
$DQ_{lymphocyte}$ & 0.675 (0.656--0.698) & 0.713 (0.691--0.734) \\
\hline
$DQ_{connective}$ & 0.566 (0.546--0.584) & 0.583 (0.565--0.602) \\
\hline
$DQ_{dead}$ & 0.410 (0.361--0.465) & 0.435 (0.306--0.561) \\
\hline
$DQ_{epithelial}$ & 0.668 (0.639--0.694) & 0.673 (0.644--0.702) \\
\hline
$DQ_{macrophage}$ & 0.657 (0.583--0.727) & 0.615 (0.531--0.703) \\
\hline
$DQ_{neutrofil}$ & 0.691 (0.625--0.753) & 0.729 (0.679--0.778) \\
\hline
\end{tabular}
%
\end{table}

\clearpage

\subsection{\label{chap:S7}QuPath integration method}
We adopt an integration strategy leveraging the paquo \cite{Bayer_AG} library, a Python package that enables direct interaction with QuPath’s internal API, thereby facilitating seamless data exchange without intermediate conversion steps. The data processing pipeline (\hyperref[fig:S7]{Appendix Figure S7}) begins with the acquisition of WSIs and their associated annotations from QuPath, which are represented as Shapely \cite{Gillies_Wel_etal._2024} polygons. Utilizing paquo, we directly read, create, and modify these annotations and detections within a QuPath project in the Python environment. Images are then cropped using these polygons and processed by cell segmentation and classification models employing standard vision processing toolkits such as OpenCV, pyvips, and PyTorch. Additionally, QuPath employs Groovy scripts to initiate a Python process that starts the entire pipeline from QuPath graphical interface: fetching polygons, extracting images from them, and running deep learning model inference on the cropped images. 
The results are returned to QuPath, leveraging paquo's Python bindings to manipulate QuPath data while minimizing the computational overhead typically associated with cross-environment communication.

\counterwithin{figure}{subsection}
\renewcommand{\thefigure}{S\arabic{subsection}}

\begin{figure}[h!]
    \centering
    \includegraphics[width=\textwidth]{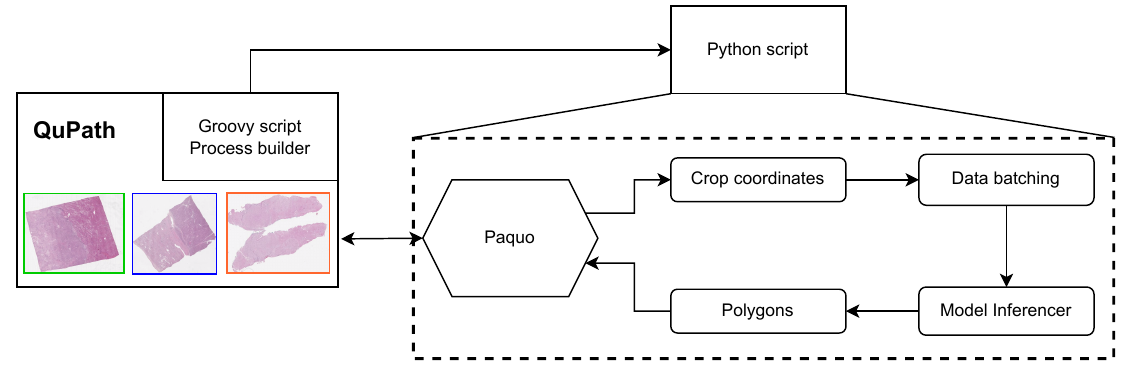}
    \caption{QuPath integration workflow using Python environment}
    \label{fig:S7}
\end{figure}

Compared to traditional workflows that involve exporting annotations as GeoJSON, classifying them in Python, and reimporting them into QuPath, our approach offers several advantages. We eliminate the need to switch between programming languages, providing a cohesive and streamlined development process entirely within QuPath software and removing the necessity to use other tools. Meanwhile, we avoid storing annotations as intermediate JSON files unless required for external use or archiving. By conducting the entire inference and post-processing workflow within the Python environment, we leverage the power and flexibility of Python libraries for image processing and machine learning. This approach also enables adjustments to any set of labels and models, thereby improving its applicability.


The distilled model and QuPath integration code are packaged into a Docker container, enabling streamlined execution with the Docker engine. Detailed integration code and deployment instructions can be found in the GitHub repository \cite{Shvetsov_2025b}.

Despite these benefits, we acknowledge that the paquo library is a proof‑of‑concept project in its early development stage and has not been tested across all versions of QuPath.

\clearpage

\subsection{\label{chap:S8}Data and code availability statement}
All datasets, models, and code used in this study are publicly available and can be obtained from the repositories listed below. 
The PanNuke \cite{Gamper_Koohbanani_etal._2019} and MoNuSAC \cite{Verma_Kumar_etal._2021} datasets are publicly accessible, and download information along with detailed descriptions can be found in their respective articles. Preprocessing scripts for PanNuke and MoNuSAC data, as well as individual cell extraction scripts, are available on GitHub \cite{Shvetsov_2025a}. The H-Optimus foundation model used in our experiments can be downloaded from the HuggingFace repository \cite{hoptimus2024}, and model information is available on GitHub \cite{Saillard_Jenatton_etal._2024}. In addition, the integration code for QuPath and the distilled model packaged in a Docker container are provided in the repository \cite{Shvetsov_2025b}, and paquo Python library is available from the authors GitHub repository \cite{Bayer_AG}.
\clearpage

\end{document}